\documentclass{article}
\usepackage{spconf,amsmath,graphicx}
\usepackage{setspace}

\usepackage{cite}
\usepackage{amsmath,amssymb,amsfonts}
\usepackage{algorithmic}
\usepackage{graphicx}
\usepackage{textcomp}
\usepackage{xcolor}
\usepackage{booktabs}
\usepackage{multirow}

\usepackage{amssymb} 
\usepackage{amsmath}
\usepackage{etoolbox}
\usepackage{amssymb}
\usepackage{graphicx}

\usepackage{algorithm}
\newcommand{\mycomment}[1]{}
 \usepackage[bb=boondox]{mathalfa}

\usepackage{algorithmic}

\newtheorem{remark}{\textbf{Remark}}

\usepackage[none]{hyphenat}  
\newcommand{\bld}[1]{\boldsymbol{#1}}
\title{Matrix Factorization in Tropical and Mixed Tropical-Linear Algebras}
\name{Ioannis Kordonis$^1$,~~Emmanouil Theodosis$^2$,~~George Retsinas$^1$,~~Petros Maragos$^1$\thanks{ The research project was supported by the Hellenic Foundation for Research and Innovation (H.F.R.I.) under the ``2nd Call for H.F.R.I. Research Projects to support Faculty Members \&  Researchers''  (Project Number:2656, Acronym: TROGEMAL). }}
\address{\textit{$^1$School of Electrical and Computer Engineering}  
\textit{National Technical University of Athens}, Greece\\\textit{$^2$School of Engineering and Applied Sciences}  
\textit{Harvard University
Cambridge, MA 02138}\\
 kordonis@central.ntua.gr,~~etheodosis@seas.harvard.edu,~~gretsinas@central.ntua.gr ,~~maragos@cs.ntua.gr}
\begin{document}
\maketitle
\begin{abstract}
 Matrix Factorization (MF) has found numerous applications in Machine Learning and Data Mining, including collaborative filtering recommendation systems, dimensionality reduction, data visualization, and community detection. Motivated by the recent successes of tropical algebra and geometry in machine learning, we investigate two problems involving matrix factorization over the tropical algebra. For the first problem, Tropical Matrix Factorization (TMF), which has been studied already in the literature, we propose an improved algorithm that avoids many of the local optima. The second formulation considers the approximate decomposition of a given matrix into the product of three matrices where a usual matrix product is followed by a tropical product. This formulation has a very interesting interpretation in terms of the learning of the utility functions of multiple users. We also present numerical results illustrating the effectiveness of the proposed algorithms, as well as an application to recommendation systems with promising results. \sloppy
\end{abstract}

\section{Introduction}

Tropical geometry is a research field combining ideas and methods from max-plus algebra (e.g., \cite{butkovivc2010max}) with algebraic geometry (see for example \cite{maclagan2009introduction}).
In the last few years, there is a developing interest in the application of tropical geometric ideas and tools to machine learning problems. Some of the applications include the analysis and simplification of piece-wise linear neural networks and the modeling  of graphical statistical models.  For a review and some recent results see \cite{maragos2021tropical}.  
\sloppy

This paper proposes some ideas and algorithms for matrix factorization over the tropical algebra and over mixed tropical/linear algebras. 
Matrix Factorization (MF) is a classical topic in Machine Learning and Data mining and MF techniques (e.g. low-rank or nonnegative MF) have found numerous and diverse applications, such as collaborative filtering, dimensionality reduction, data visualization, community detection, blind source separation,  and knowledge discovery, to name a few \cite{du2023matrix}. 

The contribution of this work is twofold. First, we propose some simple algorithms for Tropical Matrix Factorization (TMF) problem that manage to avoid a large number of locally optimal solutions and compare favorably with algorithms from the literature. Second, we introduce a new matrix factorization problem, that involves approximating a given matrix as a usual product of two matrices, followed by a tropical product with a third matrix. We refer to this problem as the Tropical Compression (TC) problem. This formulation has an interesting interpretation in terms of learning the  utility function of multiple users. Particularly,  utility functions are usually modeled as concave functions of their arguments (e.g. \cite{rosen1965existence}). We will see that TC formulation can be used to approximate a vector of utility functions with unknown arguments. We will also present an application of the proposed matrix factorizations in recommendation systems.

\textit{Related Work}:  There is some prior work to the TMF problem, that is to approximate a matrix as max-plus product of two matrices with given dimensions.   Early applications of TMF include the problem of state space realization of max-plus systems \cite{de1997matrix}. The exact formulation of TMF can be reduced to an Extended Linear Complementarity Problem (ELCP) \cite{de1995minimal}. ELCPs also describe the solution of sets of
tropical polynomial equations \cite{de1995extended}. Unfortunately, the general TMF problem is NP-hard \cite{shitov2014complexity}. An approximate technique for TMF was introduced in \cite{de1997matrix}. The algorithm was extended in \cite{omanovic2020application,omanovic2021sparse,omanovic2022faststmf}, and some applications in data mining were presented. A closely related algorithm was proposed in \cite{hook2017min}, for approximating symmetric matrices as the max-plus product of a matrix with its transpose.
 Algorithms for the related problem of approximate sub-tropical matrix factorization, i.e., matrix factorization over the max-product semi-ring were proposed in \cite{karaev2016capricorn,karaev2016cancer,karaev2019algorithms}. For a  review of  several matrix factorization formulations over non-standard algebras see \cite{karaev2019matrix}.

\section{Preliminaries}
In this section, we introduce some basic notions of max-plus or tropical algebra. The underlying space is $\mathbb R_{\text{max}} = \mathbb R\cup \{-\infty\}$. This set is equipped with two binary operations $\vee$ and $+$, where $x \vee y=\max(x,y)$ and $+$ is the usual scalar addition. In this space, maximization has the role of the usual addition and addition the role of usual multiplication. We also consider the vector space $ \mathbb R^p_{\max}$  where the internal operation $ \bld x \vee \bld y$  is defined entry-wise, i.e.,  $ [\bld x \vee \bld y]_i=\max(x_i, y_i)$ and the external operation $\lambda+\bld x$, for $\lambda\in \mathbb R_\text{max}, \bld x\in\mathbb R_{\text{max}}^p$, is defined as $[\lambda+\bld x]_i=\lambda+x_i$. 

For a matrix $\bld A \in\mathbb R_{\text{max}}^{m\times p}$ and a vector $\bld x\in \mathbb R_{\text{max}}^p$, we define the tropical matrix-vector multiplication as
\begin{equation}
[\bld A\boxplus\bld x]_i=\max_j(A_{ij}+x_j).
\end{equation}
Similarly, for matrices  $\bld A\in\mathbb R_{\text{max}}^{m\times p}$ and $\bld B\in\mathbb R_{\text{max}}^{p\times n}$, we define the tropical matrix multiplication as
\begin{equation}
[\bld A\boxplus\bld B]_{ij}=\max_l(A_{il}+B_{lj}).
\end{equation}

 Tropical polynomials are polynomials in the max-plus algebra.  A  tropical polynomial function 
 $p:\mathbb R^n\rightarrow \mathbb R_{\max}$
  is defined as
\begin{align}
p(\bld x)  = \bigvee_{i=1}^{m_p} (a_i+\bld b_i^T \bld x ),\label{t_pol_def}
\end{align}
where  
 $\bld b_i\in \mathbb R^n$, $a_i\in \mathbb R_{\text{max}}$. 
A vector of tropical polynomials is called a tropical map. Observe that a tropical map can be expressed in the form $\bld A\boxplus(\bld B \bld x)$, for appropriate matrices $\bld A,\bld B$.

For a matrix $\bld A$ the Frobenius norm is given by $\|\bld A\|_F=\sqrt{\sum_{i,j} a_{ij}^2}$.
Finally, we use $\mathbb 1$ to describe an indicator function, i.e., $\mathbb 1_{i=j}=1$ if $i=j$ and zero otherwise. 

\section{Tropical Matrix Factorization }

Assume that $\bld Y$ is an $n\times p$ matrix. The approximate Tropical Matrix Factorization problem is to find $n\times r$ and $r\times p$ matrices $\bld A,\bld B$, with given $r<\min(n,p)$ that solve the optimization problem
\begin{align}
\begin{aligned}
& \underset{\bld  A,\bld B}{\text{minimize}}
& & \| \bld Y-\bld A\boxplus \bld B\|_F^2,
\end{aligned}
\label{Trop_low_Rank_prob}
\end{align}
where $\|\cdot\|_F$ is the Frobenius norm\footnote{
We could also call the above problem as the Tropical Low Rank matrix approximation problem. However, tropical rank has at least three non-equivalent definitions (see for example \cite{maclagan2009introduction}). This formulation corresponds to the `Barvinok rank'. However, to avoid confusion we call it the TMF problem.}

We start with a simple Gradient Descent (GD) formulation for the above problem. Observe that the function $$\bld f(\bld A,\bld B)=\bld A\boxplus \bld B$$ 
is piecewise linear, and in the generic case, each entry of $[\bld A\boxplus \bld B]_{ij}$ depends on a single pair maximizing entries of $\bld A,\bld B$. 

Thus, GD takes the form
\begin{align}
\label{Matrix_Factor_Alg1}
\pi(i,j) &\leftarrow\underset{l}{\text{argmax}}\{ A_{il}+ B_{lj}\},    \\\label{Matrix_Factor_Alg2}
{ A_{il}}&\leftarrow { A_{il}}-\alpha\sum_j (A_{il}+B_{lj}-Y_{ij}){\mathbb 1}_{l=\pi_k(i,j)}\\
{ B_{lj}}&\leftarrow { B_{lj}}-\alpha\sum_i (A_{il}+B_{lj}-Y_{ij}){\mathbb 1}_{l=\pi_k(i,j)}\label{Matrix_Factor_Alg3}
\end{align}
where $\alpha$ is the step-size. In case of many maximizers in \eqref{Matrix_Factor_Alg1}, assume that one is chosen at random.

In this problem, there is a large number of local minima and stationary points. The partial derivatives with respect to all $A_{il}$ such that ${\mathbb 1}_{l=\pi(i,j)}=0$ for all $j$, are zero. Thus, if the value of $A_{il}$ is very small, the partial derivative will be always zero and local search would not be able to change it. We call the entries $A_{il}$ of matrix $\bld A $ that do not contribute to any part of $\bld A \boxplus \bld B$ \textit{ineffective}.

We then propose a  simple modification of the gradient descent scheme to mitigate this issue
\begin{align}
{ A_{il}}&\leftarrow { A_{il}}-\alpha\sum_j (A_{il}+B_{lj}-Y_{ij})s_{i,l,j}\label{Modif1}\\\label{Modif2}
{ B_{lj}}&\leftarrow { B_{lj}}-\alpha\sum_i (A_{il}+B_{lj}-Y_{ij})s_{i,l,j},
\end{align}
where $s_{i,l,j}=1$ if ${l=\pi_k(i,j)}$ and $\epsilon_k$ otherwise. We choose $\epsilon_k$ to be small positive constants. The idea behind this modification is that for all the ineffective entries of matrices   $\bld A , \bld B$ change and thus have the opportunity in the next iteration to attain the maximum in \eqref{Matrix_Factor_Alg1}.  Note that similar optimization ideas were used in the context of  neural network pruning in \cite{retsinas2020weight}. We call this method Gradient Decent with Multiplicative Noise  (GDMN). We will also study a closely related modification, where we just add a stochastic value $\varepsilon_k$  in \eqref{Matrix_Factor_Alg2}, \eqref{Matrix_Factor_Alg3}, which we call Gradient Decent with Additive Noise (GDAN). 
As we shall see in the numerical section, these modifications are surprisingly effective for avoiding bad local minima.

We then shift our attention to the case where $\bld Y$ is partially specified. That is, we do not have access to all the entries $Y_{ij}$ but only for a subset $\mathcal O \subset \{1,..,n\}\times \{1,..,p\}$. Then, problem 	
\eqref{Trop_low_Rank_prob} becomes
\begin{align}
\begin{aligned}
& \underset{\bld  A,\bld B}{\text{minimize}}
& & \| Z_{\mathcal O}\circ(\bld Y-\bld A\boxplus \bld B)\|_F^2,
\end{aligned}
\end{align}
where $Z_{\mathcal O}$ is an $n\times p $ matrix with ones in the entries $(i,j)\in \mathcal O $ and zeros elsewhere, and `$\circ$' stands for the Hadamard (element-wise) product. 

In this case \eqref{Modif1}, \eqref{Modif2} become 
\begin{align*}
{ A_{il}}&\leftarrow { A_{il}}-\alpha\sum_{j:(i,j)\in \mathcal O} (A_{il}+B_{lj}-Y_{ij})s_{i,l,j}\\
{ B_{lj}}&\leftarrow { B_{lj}}-\alpha\sum_{i:(i,j)\in \mathcal O} (A_{il}+B_{lj}-Y_{ij}) s_{i,l,j},
\end{align*}

\mycomment{
\begin{remark}
We choose $\epsilon_k$ empirically, but let us comment on the scaling of this quantity, with respect to the number of measured entries of $ A$. Assume that after many steps, the quantities  $\sum_{il} B_{il}$, $\sum_{il} C_{lj}$  have reached an (approximately) steady state value. 
 We then have
 $$\alpha\sum_{i,l}\sum_{j:(i,j)\in \mathcal O} (B_{il}+C_{lj}-A_{ij}){\mathbb 1}_{l=\pi (i,j)}\simeq\epsilon_k$$
\end{remark}

}

\section{The Tropical Compression Problem}
We first define the Tropical Compression (TC) problem.  
Assume that $\bld y_1,\dots,\bld y_N$ are datapoints in $\mathbb R^n$, with $N\geq n$. The tropical compression problem is to find a description of the given dataset as the output of a tropical map. That is, we search for datapoints $\bld x_1,\dots,\bld x_N$ in $\mathbb R^p$ with $p<n$ and matrices $\bld B\in \mathbb R^{m\times p} $ and $\bld A\in \mathbb R_{\max}^{n\times m}$ that 
solve the following problem 
\begin{align}
 \underset{\bld A,\bld B,\bld X}{\text{minimize}}
 \| \bld Y-\bld A\boxplus (\bld B \bld X)\|^2_F,
\label{Trop_Compression_prob}
\end{align}
where $\bld  X=[\bld x_1~\dots,\bld x_N]$, $\bld  Y=[\bld y_1~\dots,\bld y_N]$, and $\bld B= [\bld b_1^T~\dots~\bld b_m^T]^T$.

 We then present a motivating example. Assume that there is a set of $n$ persons and a set of $N$ items and that the preference of each person towards an item is described by a utility function.
Each item has several features and the utility of each user if they receive that item is a piece-wise linear concave function of its features\footnote{Let us note that utility functions are very often modeled as concave functions (e.g. \cite{rosen1965existence}). An intuitive reason for this choice is the principle of diminishing marginal utility. Furthermore, piece-wise linear concave functions can approximate arbitrarily well any concave function.}. 
Assume also that the features of each item $i$ are described by an unknown $p-$dimensional vector $\bld x_i$.  
 
 If $\bar {\bld Y}$ is the matrix describing the utility of each person from each item, then  $\bar Y_{ij} $ can be written as 
$$\bar Y_{ij} = \min(-\bld b_{1}^T\bld x_j-a_{i,1},\dots,-\bld b_{m}^T\bld x_j-a_{i,m}), $$
where $\bld x_j $ is the vector of characteristics of  object $j$, and $-\bld b_{l}$'s the slopes of the piecewise linear utility function. 
Then, $\bld Y=-\bar{\bld Y}$ can be written as 
$$\bld Y=\bld A\boxplus(\bld B \bld X ),$$
for appropriate matrices $\bld A,\bld B$. Particularly, $\bld B$ contains as rows the slopes of all the different users\footnote{In case where the utility function of some user $i$ does not include a slope $\bld b_l$, then $a_{il}=-\infty$.}. In the case where both the features $\bld x_j$  of the objects and the slopes $\bld b_{i,l}$ are unknown, the description of $\bld Y$ reduces to a tropical compression problem. 

 \mycomment{
\begin{remark}
This formulation, which uses shared slopes can help describe the data more compactly. 
A different formulation, without slope sharing, is described in Subsection \ref{Independent_Slopes}
\end{remark}
}

\subsection{A Numerical Algorithm for the TC Problem}
Let us transform \eqref{Trop_Compression_prob} into 
 \begin{align}
\begin{aligned}
& \underset{\bld  A\in \mathbb R_{\max}^{n\times m},\bld C \in \mathbb R^{m\times N}}{\text{minimize}}
& & \| \bld Y-\bld A\boxplus \bld C\|^2_F,\\
&\text{subj. to} && \text{rank}(\bld C)\leq p
\end{aligned}.
\label{Trop_Compression_prob2}
\end{align}
It is not difficult to see that a solution $\bld {A,B,X}$ of \eqref{Trop_Compression_prob} corresponds to a solution  $\bld {A}, \bld C$ of \eqref{Trop_Compression_prob2}, with $\bld C= \bld {BX}$. On the other hand, for a solution $\bld {A}, \bld C$ of \eqref{Trop_Compression_prob2}, we can perform a rank factorization on matrix on matrix $\bld C= \bld {B'X'}$ where $\bld B'\in \mathbb R^{m\times p'},\bld X'\in\mathbb R^{p'\times N},$ and $p' =rank(\bld C)\leq p$. By adding an appropriate number of zero columns in $\bld B'$ and zero rows in $\bld X'$, we obtain a set of matrices $\bld {A,B,X}$ solving  \eqref{Trop_Compression_prob}.

Based on formulation \eqref{Trop_Compression_prob2}, we propose a  projected gradient descent type algorithm.  The constraint set is non-convex. However, the projection on the set of rank-$p$ matrices can be easily performed using singular value decomposition (see e.g. \cite{jain2017non}). The projected version of \eqref{Matrix_Factor_Alg1}-\eqref{Matrix_Factor_Alg3}
becomes
\begin{align}
\pi(i,j) &\leftarrow\underset{l}{\text{argmax}}\{ A_{il}+ C_{lj}\},    \label{Alg_TC1}\\
{ A_{i,l}}&\leftarrow{ A_{i,l}}-\alpha\sum_j (A_{il}+C_{lj}-Y_{ij})s_{i,l,j},\\
{ \tilde C_{i,l}}&\leftarrow  { C_{l,j}}-\alpha\sum_i (A_{il}+C_{lj}-Y_{ij})s_{i,l,j},\label{Alg_TC3}\\
\bld C & \leftarrow \Pi_{\text{rank }\leq p}\left(\bld {\tilde C}\right)\label{Alg_TC4}
\end{align}
 where $  \Pi_{\text{rank }\leq p}$ is the projection onto the set of matrices with rank less than or equal to $p$.

\begin{remark}
Let us note that if $\bld X$ is known and treated as input, and $n=1$ the problem reduces to a tropical regression problem \cite{maragos2020multivariate}.
\end{remark}

 \mycomment{
\subsection{Tropical Compression with Independent Slopes} 
\label{Independent_Slopes}
A closely related TC problem is the following: given an $n\times N$ matrix $\bld Y$, find matrices $\bld B_1,\dots,\bld B_m\in \mathbb R^{n\times p} $ and $\bld X\in \mathbb R^{p\times N}$  that solve the problem 
\begin{align}
 \underset{\bld B_1,\dots,\bld B_m,\bld X}{\text{minimize}}
 \| \bld Y-(\bld B_1 \bld X)\vee\dots\vee (\bld B_m \bld X) \|^2_F,
\label{Trop_Compression_prob_AltForm}
\end{align}
where the last row of $\bld X$ consists of ones.  The interpretation is similar with \eqref{Trop_Compression_prob}. Particularly, $$Y_{ij} = \max_l(\bld b_{l,i}^{T}  \bld x_j)= \max_l(\tilde{\bld b}_{l,i}^T  \tilde{\bld x}_j+\tilde a_{l,i}),$$
where $\bld b_{l,i}^T $ it the i-th row of $\bld B_l$, $\bld x_j$ is the j-th column of $\bld X$, and $\tilde{\bld b}_{l,i}^T $, $\tilde{\bld x}_j$ represent the vectors of the first $p-1$ entries of  ${\bld b}_{l,i}^T $, ${\bld x}_j$ and $\tilde a_{l,i}$ the last entry of ${\bld b}_{l,i}^T$. 

\begin{remark}
Let us note that if $\bld X$ is known and treated as input, the problem reduces to $n$ independent tropical regression problems \cite{maragos2020multivariate}.
\end{remark}
}

\section{Numerical Examples}
\subsection{Synthetic Data}

At first we implement the proposed schemes in Python package CuPy (a version of NumPy that  allows for GPU acceleration).

\subsubsection{Tropical Matrix Factorization}
We first present an example that illustrates the usefulness of the proposed modifications GDMN and GDAN. 
We chose a matrix $\bld Y$ given as 
\begin{equation}
\bld Y = \bar{\bld A}\boxplus \bar{\bld B} +a\bld R,
\end{equation}
where $\bar{\bld A},\bar{\bld B},\bld R$ are $10\times 5$, $5\times 11$, and $10\times 11$ matrices the entries of which are chosen as i.i.d. random variables following the uniform distribution on $[0,1]$, and $a=0.1$.  

\begin{figure}
\centering
	\includegraphics[width=0.5\textwidth]{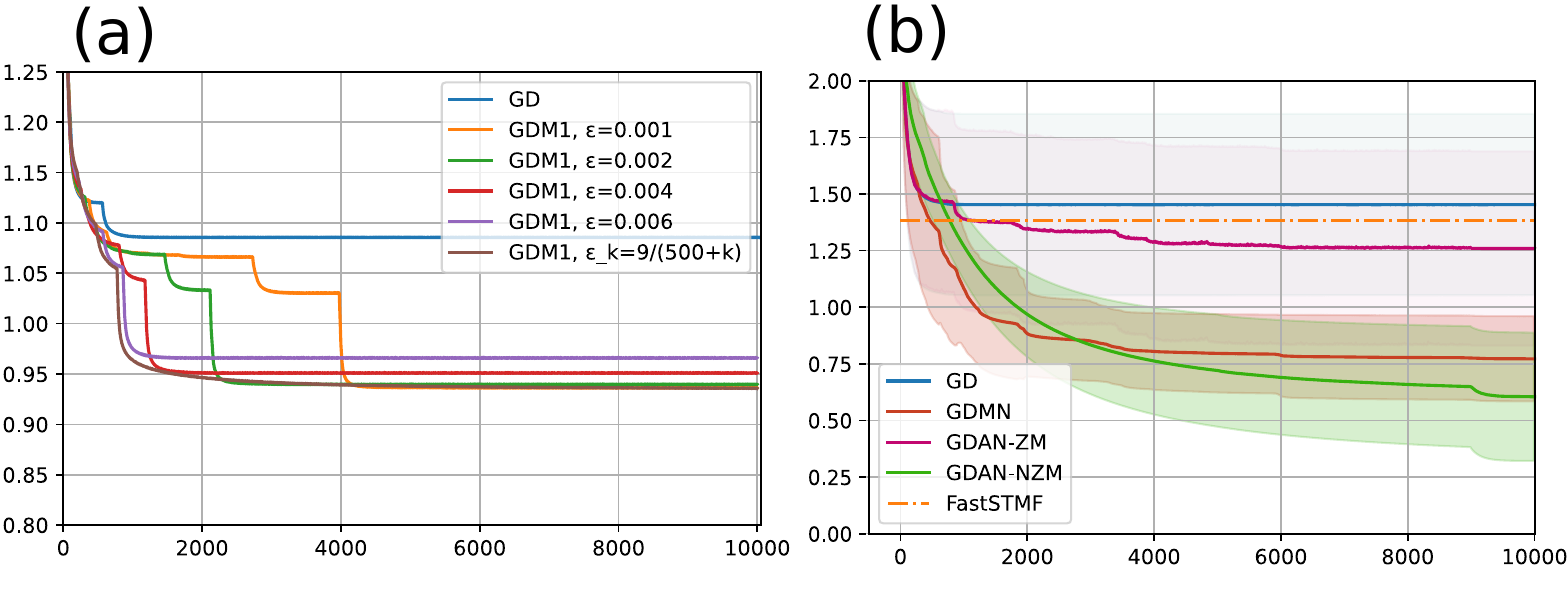}
\caption{\textit{(a) The approximation error norm for Algorithm \eqref{Modif1}-\eqref{Modif2}. (b) The comparison of the GD scheme and the proposed modifications, with FastSTMF. The results for (b) have been averaged over 10 runs. FastSTMF converges very fast (in $2$ to $5$ steps) but does not improve further. Since it uses a different kind of iteration, we included only its final value. }}
\label{MatFactorCurves}
\end{figure}

Figure \ref{MatFactorCurves}.a compares the norm of the error $\| \bld Y -\bld A\boxplus\bld B\|_F$, where the matrices $\bld A, \bld B$ are computed using Algorithm  \eqref{Modif1}-\eqref{Modif2}, with different values of $\epsilon$ (recall that $\epsilon$ represents the contribution of non-maximizing entries to the algorithm). The modification allows the algorithm to overcome some local optima. Observe that there is a trade-off between convergence speed and quality of solution.  With a large value of $\epsilon$, we have faster convergence to a worse solution. Additionally, we used a diminishing scheme for $\epsilon$ in the form $\epsilon_k=9/(500+k)$, where $k$ is the iteration count. All the results are normalized, that is, we divide the error with the norm $a\|R\|_F$. Figure \ref{MatFactorCurves}.b compares the modified versions of GD.

We then compare the proposed algorithms with the FastSTMF algorithm from \cite{omanovic2022faststmf}. We factorize multiple $10\times 11$ matrices with $r=5$. To have a fair comparison, we use as an initial estimate in both algorithms the matrices proposed in  \cite{omanovic2022faststmf}. Table 1 shows the matrix factorization error of FastSTMF  and compares it with the error of the GD and the proposed variations. We have two implementations of GDAN, with zero mean (GDAN-ZM) and non-zero mean (GDAN-NZM). GDAN-NZM has the best performance between the variants examined. This is probably because it promotes competition between the different non-maximizing entries of matrices $\bld A,\bld B$.

\begin{table}[h]
    \centering
\scalebox{0.97}{    \begin{tabular}{llll}
        \toprule
        \multirow{1}{*}{\textbf{Algorithm}} & \multicolumn{3}{c}{Parameter $a$} \\
        \cmidrule{2-3} \cmidrule{4-4} 
        {} & $~~~~~~0.01$ & $~~~~~~0.1$& $~~~~~~0.5$\\
        \midrule
        FastSTMF   \cite{omanovic2022faststmf}   & $11.2\pm 4.9 $    &   $1.38\pm 0.52 $                 & $  .52\pm  .06 $   \\
        GD             & $13.2\pm 3.5 $    &  $ 1.45 \pm 0.40  $                  &  $  .43 \pm .04  $  \\
        GDMN         & $06.3\pm 4.1 $  &   $ 0.77\pm 0.19  $                    &    $  .39 \pm .04  $ \\
        GDAN-ZM     & $11.4\pm2.6$     &  $1.26 \pm 0.43  $                  &    $  .42 \pm  .04 $\\
        GDAN-NZM   & $\bld{4.5\pm 2.1}  $  &  $\bld{ 0.60\pm 0.28 }$   &   $   \bld{.34\pm  .05} $ \\
        \bottomrule
    \end{tabular}}
\caption{Comparison of the Algorithms: The normalized Frobenius error, for various values of $a$. }
\end{table}

 
\subsection{Real Data}
\subsubsection{Movielens 100k Dataset}
We use the Movielens 100k Dataset \cite{harper2015movielens}, consisting of the ratings of $943$ users to $1682$ movies. There are in total $100000$ ratings. Here we use the implicit feedback formulation. That is, we consider a matrix $\bld Y$ with a value of $-1$ if the person has watched a movie and $+1$ if they haven't.

We then use a factorization of matrix $\bld Y$. We split the data into $80\%$ training, $10\%$ validation, and $10\%$ test, and apply a stochastic version of GD, and early stopping. We use two metrics, the RMS error and the  Hit Rate at 10 (HR@10)\footnote{HR@10 is defined as follows. For each user, form a list of $101$ items choosing randomly from the test set $1$ positive and $100$ negative items. Then, count the number of users for which the positive item is ranked among the first $10$ of the list  \cite{anelli2021reenvisioning}.}.

The best approximation comes for an intermediate dimension $r=35$ and has an RMS error equal to $0.396$ in the test set and HR@10 is $0.755$.
We then consider the TC formulation of the problem \eqref{Trop_Compression_prob}, with $m=40$ and $p=25$. Using the modified projected gradient descent algorithm \eqref{Alg_TC1}-\eqref{Alg_TC4} we get an   RMS error equal to $0.391$ and HR@10 equal to $0.77$. Compared to the TMF formulation, TC performs slightly better. It has also a smaller number of parameters and an intuitive interpretation.

\vspace{-5pt}

\subsubsection{Movielens 1M Dataset}

We then turn to a larger dataset, Movielens 1M, with $1$ million ratings from $6000$ users on $4000$ movies. We  formulate matrix $\bld Y$, as in the previous subsection. Then, using the same train/validation/test split, we compute an approximate tropical factorization for matrix $\bld Y$, with $r=40$. Then, the RMS error becomes $0.328$ and the HR@10 becomes $0.742$. For comparison, a carefully optimized and regularized linear factorization  gives HR@10  equal to $0.731$ \cite{anelli2021reenvisioning}. For a TC formulation with $m=100$ and $p=35$,  the RMS error becomes  $ 0.327$ and  HR@10 becomes $ 0.743$.


\section{Conclusion and Future Work}
This paper formulates two matrix factorization problems, over the tropical algebra and over mixed linear tropical algebras respectively. For the first problem, we proposed some variations of Gradient Descent that lead to improved performance and compare favorably with an algorithm from the literature. For the second problem which, has an interesting interpretation, in terms of learning the utility function of a set of users, we proposed a non-convex projection gradient descent  algorithm. The proposed algorithms were applied to a recommendation problem, using datasets MovieLens 100k and 1M, with promising results.
 
Some interesting directions  for further research are the  use of appropriate regularization techniques and the study of sparse approximate solutions for the TMF problem. \mycomment{Additionally, it would be interesting to explore  other applications of TMF, including on data visualization. Finally, a deeper link between TMF, TC, and tropical regression can be investigated. }

\bibliographystyle{IEEETran}
\bibliography{refs}

\end{document}